\documentclass[runningheads]{llncs}
\usepackage[T1]{fontenc}
\usepackage{array}
\usepackage{graphicx}
\usepackage{amsmath, amssymb}
\usepackage{multirow}
\usepackage{tabularx}
\usepackage{rotating}
\usepackage{booktabs}
\usepackage{subcaption}
\usepackage{makecell}
\usepackage{hyperref}
\usepackage{longtable}

\usepackage{xcolor}    % for colors
\usepackage{soulutf8}  % for proper highlighting with UTF-8

\newif\ifreviewmode
\ifreviewmode
    \newcommand{\review}[1]{\textcolor{blue}{#1}}
\else
    \newcommand{\review}[1]{#1}
\fi

\newcommand{\dt}{\Delta t}
\newcommand{\R}{\mathbb{R}}

\begin{document}
\title{The Impact of Temporal Granularity on Socio-Demographic Inference from Household Load Profiles}
\author{Dejan Radovanovic\inst{1,2} \and
Maximilian Schirl\inst{1} \and
Andreas Unterweger\inst{1,2} \and
Günther Eibl\inst{1}}
\authorrunning{Radovanovic et al.}
\institute{Center for Secure Energy Informatics, Salzburg University of Applied Sciences, Puch bei Hallein, Austria \and
Paris Lodron University of Salzburg, Salzburg, Austria}
\maketitle              % typeset the header of the contribution
\begin{abstract}
Smart meter data can reveal sensitive socio-demographic characteristics of households, raising privacy concerns. While this risk has been demonstrated at fixed granularities, the role of temporal resolution in shaping inference performance remains insufficiently explored. This paper addresses this gap by analyzing how load profiles with granularities from 15 minutes to 7 days affect the predictability of eight socio-demographic attributes in a dataset of 1,589 households over one year. We introduce an evaluation framework where classifiers are trained on year-round data but tested on arbitrary weeks, forcing generalization across seasonal and weekly variations.

Our results show three main findings. First, while coarsening granularity reduces predictive accuracy, two plateaus emerge: performance is stable between 15 minutes and 1 hour, and again between 1 and 7 days. This reveals opportunities for data minimization without sacrificing utility. Second, interpretable handcrafted and ts-fresh features remain competitive with CNN-based autoencoder embeddings, while XGBoost consistently outperforms alternative classifiers. Third, feature importance analysis highlights differences between static and dynamic attributes: dwelling size can be inferred even from coarse data, whereas swimming pool usage requires fine-grained temporal signals.

Overall, our study provides new insights into the privacy–utility trade-off in smart metering, showing how temporal resolution, feature extraction, and classifier choice jointly influence socio-demographic inference.

\end{abstract}
\section{Introduction}
\label{sec:Intro}
\review{Smart metering enables the collection of high-resolution electricity consumption data, offering utilities and consumers detailed insights into household energy usage patterns. Such load profiles support diverse applications, including short-term load forecasting, demand response programs, dynamic pricing schemes, and improved energy efficiency, ultimately contributing to reduced energy waste and cost savings~\cite{darby2010smart,Weranga13a}.}

\review{However, the utility of load profiles depends critically on their temporal resolution. Coarse monthly consumption values are sufficient for billing, yet they cannot capture daily or hourly variations required for accurate household-level forecasting. At weekly granularity, consumption peaks and appliance-specific patterns are smoothed out, rendering short-term predictions ineffective. This illustrates the fundamental privacy–utility trade-off: coarser resolutions increase privacy by reducing behavioral detail but simultaneously limit the utility of the data for advanced analytics.}

\review{The widespread deployment of smart meters has raised significant privacy concerns. Fine-grained consumption data allow inferences about household behavior, such as appliance usage~\cite{Chen13a,Kleiminger13a,Abreu12a} and occupancy schedules~\cite{Kim11d,Kolter12a,Fan13a}. More critically, because electricity usage is often correlated with socio-demo- graphic attributes, such as dwelling type, household size, or income level, load profiles can be exploited to predict sensitive household characteristics~\cite{Beckel11a,Beckel13a,Beckel14a,Hopf16a,Wang19a}. In the worst case, predictions of socio-demographics, such as whether a household has a swimming pool, sauna, or owns their home, may serve as proxies for social standing or income group. Even without linking to external data, such inferences can have discriminatory consequences: (i) insurers could adjust premiums based on inferred socio-economic status, (ii) advertisers might micro-target families, (iii) landlords or financial institutions could indirectly evaluate financial capacity, and (iv) utilities could tailor tariffs that disproportionately affect specific groups. Thus, socio-demographic inference from load profiles directly undermines household privacy.}

\review{A commonly proposed mitigation strategy is to reduce the resolution of load profiles, thereby increasing their time granularity~\cite{Efthymious10a,Eibl15a,Engel17a,Erkin13a,Finster14a}. Prior work has primarily examined this trade-off in the context of sub-minute or second-level data, typically used for non-intrusive load monitoring (NILM), where the focus lies on appliance detection~\cite{Hart92a,Zoha12a}. In contrast, the privacy implications of coarse-grained load profiles—with temporal resolutions from 15 minutes up to several days—remain insufficiently understood. This question is highly relevant in practice, as European Union regulations mandate a minimum resolution of 15 minutes for electricity data collection~\cite{EU12b}. While such resolutions balance privacy and utility at the system level, they may still allow household-level socio-demographic inference~\cite{Lisovich10a,Molina10a,Eibl15a}.}

\review{This paper addresses the following research question: How does the time granularity of coarse-grained load profiles affect the predictability of household socio-demographic characteristics? Earlier studies have shown that socio-demographics can be inferred from 30-minute profiles~\cite{Beckel13a,Beckel14a,Hopf16a} or have examined general effects of temporal resolution~\cite{Eibl15a}, but a systematic analysis across granularities from 15 minutes to 7 days is still missing~\cite{Alahakoon16,Wang18b,Ashgar17a}. To close this gap, we evaluate the inference of eight socio-demographic attributes using weekly load profiles from 1,589 suburban households observed over one year. Unlike studies that rely on fixed seasonal or annual aggregates, our framework evaluates classifiers on arbitrary weeks, requiring generalization across seasonal variation and weekly consumption dynamics.}

\review{This work extends our previously published conference paper~\cite{Radovanovic25a} with several additional contributions:}
\begin{enumerate}
    \item \review{Refined evaluation of seasonal attributes: While the set of socio-demographic characteristics remains the same as in the previous study (e.g., swimming pool, sauna, home ownership), we extend the analysis by restricting seasonal attributes to relevant time periods (e.g., swimming pool usage evaluated only during summer weeks, sauna during winter weeks).}
    \item \review{Systematic comparison of feature extraction methods: We contrast handcrafted features, ts-fresh, and CNN-based autoencoder embeddings across all granularities.}
    \item \review{Expanded classifier analysis: Multiple machine learning models are evaluated to test robustness across temporal resolutions.}
    \item \review{Feature importance and interpretability: We identify which consumption features most strongly drive socio-demographic inference.}
\end{enumerate}

\review{Finally, the privacy–utility trade-off studied here has tangible real-world implications. While coarse data (e.g., monthly billing) suffice for operational tasks, fine-grained load profiles enable the prediction of sensitive household characteristics. In worst-case scenarios, such inferences could facilitate profiling for insurance pricing, targeted advertising, or discriminatory practices. These risks underscore the importance of carefully balancing data utility with privacy preservation in future smart metering deployments.}

The remainder of this paper is structured as follows: Section~\ref{sec:Related_Work} reviews existing work on socio-demographic inference from smart meter data and the impact of time granularity. Section~\ref{sec:Problem_Definition} formally defines the classification problem under study. Section~\ref{sec:Methodology} describes our experimental setup, including dataset preprocessing, feature extraction, and classification methods. Section~\ref{sec:Results} presents the empirical results, while Section~\ref{sec:Discussion} discusses the implications and compares our findings with related approaches. Finally, Section~\ref{sec:Conclusion} concludes the paper and outlines directions for future research.

\section{Related Work}
\label{sec:Related_Work}
\review{Smart meter data has been widely used to analyze residential electricity consumption and infer household characteristics. Prior research spans domains such as energy efficiency, demand forecasting, non-intrusive load monitoring (NILM), and privacy-preserving analytics. While NILM studies demonstrate how fine-grained second-level data enables appliance-level disaggregation~\cite{Zeifman11a,Zoha12a}, our focus lies on coarse-grained data starting at 15-minute intervals, in line with European Commission guidelines~\cite{EuropeanCommission2014}.} 

\review{Early work on coarser data investigated small datasets (5-30 households) using daily or sub-daily intervals, for tasks such as anomaly detection~\cite{Verdu06a}, identifying routines~\cite{Abreu12a}, or short-term load forecasting~\cite{DeSilva11a}. Larger studies linked hourly consumption to external factors such as temperature~\cite{Birt12a}. Other research explored how socio-demographic characteristics correlate with load profiles. McLoughlin et al.~\cite{McLoughlin12a,McLoughlin15a} used the CER dataset to relate dwelling type and household composition to consumption and to cluster households into usage classes. Similarly, Kolter~\cite{Kolter11b} and Kavousian~\cite{Kavousian13a} analyzed U.S. data, showing the influence of weather and dwelling characteristics.} 

\review{A central line of work predicts socio-demographics directly from load profiles. Beckel et al.~\cite{Beckel13a,Beckel14a} used 30-minute CER data to classify attributes such as dwelling type and occupancy, while Hopf et al.~\cite{Hopf16a} improved accuracy by expanding the feature set. Viegas et al.~\cite{Viegas16a} applied fuzzy models for interpretable predictions, and Wang et al.~\cite{Wang19a} compared handcrafted and automatically extracted features. Studies closer to our dataset investigated specific appliances: Burkhart~\cite{Burkhart18a} and Ferner~\cite{Ferner19a} predicted swimming pool ownership, though both assumed year-long load profiles for training and testing. In contrast, our work evaluates \textit{arbitrary weekly snippets}, requiring classifiers to generalize across seasonal variation.} 

\review{More recent work has introduced alternative methodological advances. Lin et al.~\cite{Lin22a} applied Bayesian CNNs with synthetic oversampling to address class imbalance and quantify prediction uncertainty. Their emphasis lies on uncertainty-aware classification with half-hourly data, whereas our study systematically examines the role of temporal granularity in privacy risks. Fahim and Sillitti~\cite{Fahim19a} combined \texttt{ts-fresh} features with Random Forests on high-resolution data aggregated to 30 minutes, focusing on utility for demand-response. In contrast, we compare handcrafted, ts-fresh, and autoencoder embeddings across granularities up to 7 days, explicitly framing the task in terms of privacy implications.}  

\review{Few works explicitly study time granularity itself. Eibl and Engel~\cite{Eibl15a} showed that increasing aggregation reduces appliance detection accuracy, while Engel et al.~\cite{Engel17a} proposed multi-resolution encrypted load data for privacy protection. In contrast, our contribution is twofold: (i) we analyze socio-demographic inference from coarse-grained load profiles (15 minutes to 7 days) using a large dataset of 1,589 households, and (ii) we evaluate how random weekly profiles, rather than fixed seasonal or annual aggregates, affect classifier performance, highlighting the privacy risks of socio-demographic predictability across temporal resolutions.}  

\review{Beyond prediction accuracy, we also study \emph{feature importance across varying time granularities}, which provides additional interpretability. While prior work has mainly reported overall classification performance~\cite{Beckel14a,Hopf16a,Wang19a}, our analysis shows how the relevance of specific load-profile characteristics shifts as temporal resolution changes. This perspective offers deeper insights into how aggregation not only reduces accuracy but also alters the consumption patterns that remain informative for inferring household attributes.}  

\review{Another methodological difference concerns the temporal unit of analysis. Previous work often relied on fixed seasonal subsets (e.g., summer weeks for swimming pool detection~\cite{Ferner19a}) or entire year-long profiles~\cite{Burkhart18a}. In contrast, we design a more realistic and challenging evaluation by sampling \emph{arbitrary weekly snippets}. This setting reflects a practical threat scenario in which an adversary may only obtain limited, temporally unconstrained data, yet still attempts to infer sensitive household attributes. To our knowledge, this perspective, combined with a systematic comparison of feature extraction strategies, classifiers, and feature importance across multiple granularities, has not been addressed in prior work.}

\section{Problem Definition}
\label{sec:Problem_Definition}
\review{In a threat scenario where household electricity consumption data are disclosed together with corresponding socio-demographic attributes, such as in datasets described by~\cite{Beckel14a,Burkhart18a,Ferner19a,Radovanovic22a}, an adversary obtains weekly load profiles $w$ for households $h$ along with their socio-demographic characteristics.
Electricity consumption is observed at a given temporal resolution $\dt$. The adversary’s objective is to construct a classifier $f^{\dt}$ capable of predicting a binarized socio-demographic attribute (label) $y$ of a household from its load profile.}

\review{Once trained, the classifier $f^{\dt}$ is applied to an unseen household. Given a consumption segment $c^{\tilde{w},\dt}$ representing an arbitrary week $\tilde{w}$ of the year, the model predicts the label $y$. Formally, the task is defined as:}

\begin{equation}
f^{\dt}: \R^{n} \to {0,1}, \quad c^{\tilde{w},\dt} \mapsto y
\end{equation}

\review{This classifier can subsequently be deployed on weekly load profiles even in the absence of ground-truth socio-demographic information, thereby raising privacy concerns. The dimensionality $n$ of the consumption vector $c^{\tilde{w},\dt}$ depends on the temporal resolution: as $\dt$ becomes coarser, $n$ decreases, as illustrated in Table~\ref{tabel_time_resolution}.}

\review{The central objective of this study is to examine how time granularity $\dt$ affects the predictive performance of $f^{\dt}$. In line with prior findings, it is expected that classification accuracy deteriorates as the resolution becomes coarser.}

\section{Experimental Setup and Methodology}
\label{sec:Methodology}
\begin{figure*}[ht]
  \centering
  \includegraphics[width=\linewidth]{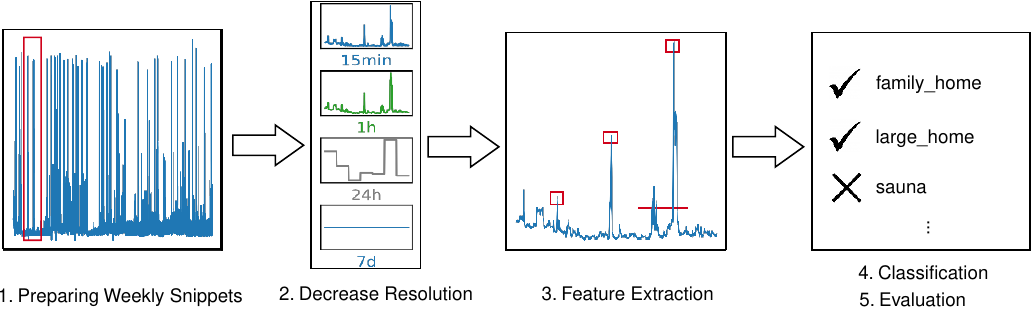}
  \caption{Methodological overview consisting of five steps: (1) preparing weekly snippets, (2) decreasing temporal resolution, (3) feature extraction, (4) classification, and (5) evaluation, as already used in~\cite{Radovanovic25a}.}
  \label{figure_methodology_pipeline}
\end{figure*}

We employ supervised machine learning techniques to assess how different temporal granularities of load profiles affect the inference of household-specific socio-demographic characteristics. The methodology, illustrated in Figure~\ref{figure_methodology_pipeline}, comprises five main stages: preparation of weekly snippets, decrease of temporal resolution, feature extraction, classification, and evaluation.

Section~\ref{subsection_dataset} introduces the input dataset, which contains one year of 15-minute load profiles and corresponding household socio-demographic attributes. The selection of target characteristics is described in Section~\ref{subsection_selection_of_labels}, and the preprocessing steps used to generate weekly snippets are detailed in Section~\ref{subsection_preprocessing}. Temporal down-sampling is addressed in Section~\ref{subsection_decrease_resolution}, followed by the computation of numerical feature representations in Section~\ref{subsection_feature_extraction}. The extracted features are then classified using off-the-shelf models (Section~\ref{subsection_classification}) and evaluated in Section~\ref{subsection_evaulation}. Finally, Section~\ref{subsection_feature_importance} analyzes the importance of individual features to better understand their contribution to the prediction of socio-demographic attributes.

\subsection{Dataset}\label{subsection_dataset}
The used dataset, PEAK Load Data, stems from a field test, which collected electricity consumption profiles of 1,589 suburban households in Upper Austria via smart meters between September 30, 2017 and October 15, 2018. 
The field test aimed at testing various incentive schemes for motivating electricity consumers to shift loads towards times of high renewable production. More information about the study and the collection of the data can be found in~\cite{Radovanovic22a}.

The acquired data contains 15-minute load profiles and household-specific socio-demographic characteristics e.g., household type, household size, household appliances and heating type. To put some of the suburban household characteristics into perspective, the following statistics are illustrated: 
The yearly average energy consumption per household is 5327 kWh, with the median being 4409 kWh and a standard deviation of 3721 kWh. The average household size is 138 (mean), 130 (median) and 58 (standard deviation) in square meters, respectively. With 2.8, 2 and 1.2 residents per household, respectively. 

\subsection{Selection of household-specific characteristics and class labels~\label{subsection_selection_of_labels}}
Table~\ref{tabel_dataset_description} summarizes all available household-specific characteristics collected during the field test, including their absolute frequencies, class imbalance ratios, and the number of positive and negative samples. For the prediction task, a subset of these characteristics has been selected as class labels based on the following criteria:
\begin{enumerate}
    \item \review{\textbf{Privacy sensitivity}: Prior research has shown that socio-demographic attributes such as household composition (e.g., family homes versus single households) and home ownership status pose a higher privacy risk compared to appliance ownership or heating types~\cite{Beckel12a}. Characteristics with a stronger potential to reveal private information are therefore prioritized.}
    \item \textbf{Comparability with prior studies}: To enable direct comparability with existing results on socio-demographic inference from energy consumption data, we select characteristics that overlap with those analyzed in~\cite{Beckel14a,Hopf16a,Wang19a}.
    \item \textbf{Class balance}: To ensure reliable classification and avoid severe data imbalance issues, we select characteristics with an imbalance ratio below a threshold of $5$. \review{This threshold ensures that the minority class represents at least $20\%$ of the samples, which is generally considered sufficient for stable and reliable classifier training without the need for additional re-balancing techniques~\cite{He09a}. A stricter threshold would have removed several privacy-relevant characteristics, while a more lenient one would have led to severely skewed class distributions, potentially biasing both performance evaluation and privacy-risk estimation.}
\end{enumerate}

\review{In addition to these selection criteria, certain attributes are seasonally restricted to reflect their relevance in practice: the label \textit{swimming\_pool} is evaluated only for summer months (May–September), when pool usage and associated electricity demand occur, while \textit{sauna} is analyzed exclusively for winter months (November–March), when heating-related consumption patterns are present. Furthermore, multi-class and numeric characteristics are simplified into binary labels to reduce complexity and ensure consistency across classification tasks. This binarization follows threshold values commonly used in prior work~\cite{Beckel13a,Wang19a}; for instance, households with more than two residents are labeled as \textit{family\_home}, and those with a living area exceeding $100$~m$^2$ as \textit{large\_home}. Since socio-demographic information is reported once per household, the same label is uniformly assigned to all $52$ weekly snippets of that household. Importantly, adopting binary classification also enhances interpretability, as SHAP values can directly quantify each feature’s contribution to revealing the presence or absence of a specific socio-demographic characteristic, whereas multi-class or numeric formulations would complicate the explanation process without offering additional privacy insights.}

The final set of labels used for prediction is highlighted in bold in Table~\ref{tabel_dataset_description}. This selection provides a balanced and privacy-relevant set of characteristics while maintaining alignment with prior research, thereby supporting robust and interpretable classification results.

\begin{table}[!ht]
\caption{Extended list of household characteristics and their positive and negative samples, sorted by imbalance ratio, building on~\cite{Radovanovic25a}.}
\label{tabel_dataset_description}
\centering
\setlength{\tabcolsep}{3pt}
\begin{tabular}{|l|c|c|c||l|c|c|c|}
\hline
\thead{Characteristic} & \thead{Pos.} & \thead{Neg.} & \thead{Ratio} &
\thead{Characteristic} & \thead{Pos.} & \thead{Neg.} & \thead{Ratio} \\
\hline
\textbf{family home} & \textbf{595} & \textbf{613} & \textbf{1.03} &
oil heating & 177 & 1031 & 5.82 \\
dryer & 713 & 495 & 1.44 &
district heating & 162 & 1046 & 6.46 \\
\textbf{heat pump} & \textbf{400} & \textbf{808} & \textbf{2.02} &
biomass heating & 160 & 1048 & 6.55 \\
\textbf{split house} & \textbf{829} & \textbf{379} & \textbf{2.19} &
heat pump water & 142 & 1066 & 7.51 \\
\textbf{large home} & \textbf{831} & \textbf{377} & \textbf{2.20} &
oil water & 102 & 1106 & 10.84 \\
deep freezer & 868 & 340 & 2.55 &
biomass water & 86 & 1122 & 13.05 \\
\textbf{pool} & \textbf{328} & \textbf{880} & \textbf{2.68} &
aquarium & 75 & 1133 & 15.11 \\
\textbf{apartment} & \textbf{283} & \textbf{925} & \textbf{3.27} &
water bed & 70 & 1138 & 16.26 \\
gas heating & 280 & 928 & 3.31 &
electric heating & 53 & 1155 & 21.79 \\
electric water & 274 & 931 & 3.40 &
air condition & 42 & 1166 & 27.76 \\
\textbf{sauna} & \textbf{256} & \textbf{952} & \textbf{3.72} &
computer & 1188 & 20 & 59.40 \\
\textbf{home owned} & \textbf{967} & \textbf{241} & \textbf{4.01} &
 & & & \\
\hline
\end{tabular}
\end{table}

\subsection{Preparing weekly snippets}\label{subsection_preprocessing}
Preprocessing included the following steps: (i) removal of households with too many missing values, (ii) grouping load profiles into weekly time snippets and (iii) selection and binarization of labels for later classification.

\review{The original dataset contains smart meter measurements for 1589 households. Due to incompleteness in some profiles—caused by disruptions, device breakdowns, or smart-meter replacements—the number of usable households is reduced to $n=1208$. A household is excluded if it lacks at least one complete week of measurements (i.e., fewer than $672$ valid 15-minute readings per week), as such profiles are unsuitable for constructing consistent weekly load snippets.}

\review{To ensure seasonal coverage and comparability across households, only the common time frame of 52 full weeks (October 2, 2017 to September 29, 2018) is considered. For each household, the yearly load profile comprising $52\cdot7=364$ days is regrouped into 52 weekly snippets as described in~\cite{Radovanovic22a}, resulting in a data matrix of size $n \times m$ with $n=1208\cdot52=62816$ household-week combinations and $m=7\cdot24\cdot4=672$ measurements per week.}

\review{Although it is possible that a household may be on vacation during certain weeks, a weekly snippet is considered a realistic and practical time frame for adversarial inference: it represents the smallest continuous period that an attacker could plausibly access and still capture behavioral patterns relevant for socio-demographic classification~\cite{Beckel13a}. Aggregating over longer periods (e.g., months or full years) would provide more stable data but assumes extended access, whereas a single week balances minimal accessibility with representativeness of household energy usage behavior.}

\subsection{Decrease of temporal resolution}\label{subsection_decrease_resolution}
\review{Before analyzing the privacy influence of different time granularities $\dt$, the weekly load profiles $w$ sampled at 15-minute intervals are downsampled to coarser resolutions. Let $h$ denote a household and $c^{w, 15\mathrm{min}}$ the weekly energy consumption snippet of household $h$ in week $w$ at the base resolution of 15 minutes. Each measurement represents the \emph{energy consumed during the respective 15-minute interval}, expressed in kilowatt-hours (kWh). 
To obtain a coarser time granularity $\dt' > 15\mathrm{min}$, we aggregate consecutive 15-minute measurements by summation:
\[
    c^{w,\dt'}_j = \sum_{i \in I_j} c^{w, 15\mathrm{min}}_i,
\]
where $I_j$ denotes the set of 15-minute indices that fall into the coarser time slot $j$. This preserves the physical meaning of the data, i.e., the total energy consumed in the larger time window. 
For example, if the meter records $0.25$~kWh and $0.30$~kWh in two consecutive 15-minute slots, the aggregated 30-minute value is $0.55$~kWh. Averaging these values would underestimate the total consumption and distort energy usage statistics.}

\review{This aggregation is applied to all target resolutions listed in Table~\ref{tabel_time_resolution}. 
For each $\dt'$, a separate data matrix 
\[
    C^{\dt'} = 
    \big[ c^{w_1,\dt'}, c^{w_2,\dt'}, \dots, c^{w_{n},\dt'} \big]^\top
\]
is generated, where $n$ denotes the number of weekly profiles across all households $h$. The number of columns $m$ decreases proportionally to the factor $\frac{\dt'}{15\mathrm{min}}$. For instance, $\dt'=1$ hour produces $m=168$ values per week, while $\dt'=24$ hours results in $m=7$ values.}

\begin{table}[!ht]
\centering
\caption{List of time granularities for decreasing temporal resolution and the resulting number of measurements $m$ for a weekly snippet.}
\label{tabel_time_resolution}
\begin{tabular}{|c|c||c|c|}
    \hline
    \thead{$\dt'$} & \thead{$m$} & \thead{$\dt'$} & \thead{$m$} \\
    \hline
    15~minutes & 672 & 12~hours & 14 \\
    30~minutes & 336 & 24~hours    & 7  \\
    1~hour     & 168 & 2~days   & 4  \\
    2~hours    & 84  & 3~days   & 3  \\
    4~hours    & 42  & 7~days   & 1  \\
    6~hours    & 28  &  &\\
    \hline
\end{tabular}
\end{table}
\review{Since subsequent classifiers $f^{\dt'}$ operate on these aggregated snippets $c^{\tilde{w},\dt'}$, temporal down-sampling is performed for all resolutions $\dt'$ to enable consistent evaluation of privacy risks across varying time granularities.}

\subsection{Feature extraction}\label{subsection_feature_extraction}
\review{Feature extraction transforms preprocessed load profiles into compact numerical representations that can be effectively used for classification~\cite{Bishop06a}. In this work, we evaluate three different approaches for extracting predictive information from weekly household consumption snippets across all time granularities $\Delta t$: (i) automatic feature generation using the ts-fresh library\footnote{\href{https://tsfresh.readthedocs.io/en/latest/}{https://tsfresh.readthedocs.io}}, (ii) handcrafted features adapted from Beckel et al.~\cite{Beckel12a,Beckel13a,Beckel14a}, and (iii) representation learning via an autoencoder architecture~\cite{Wang19a}.}

\review{The handcrafted feature set consists of 35 descriptors grouped into five categories: (i) consumption aggregates, (ii) ratios between time segments, (iii) temporal dynamics, (iv) statistical properties, and (v) the first ten principal components. These features capture a broad range of consumption behaviors, from daily totals to short-lived events. For instance, variance and peak-related features are particularly informative at fine resolutions (15 minutes), where short-lived but characteristic activities remain visible, whereas coarser resolutions emphasize cumulative consumption trends. Statistical assumptions of normality~\cite{Osborne02a,Beckel14a} are addressed by applying standard normalization, and undefined operations (e.g., ratios at coarse resolutions) are set to zero. An overview of the 35 handcrafted features and their maximum computable resolution is provided in Table2 of\cite{Radovanovic25a}.}

\review{The ts-fresh library computes a large number of statistical, temporal, and frequency-domain descriptors, including coefficients such as CWT and FFT~\cite{Christ18a}. While many of these features lack straightforward interpretability, they provide a rich feature space for comparison with other methods. To ensure fair comparison with the handcrafted set, we apply univariate feature selection using ANOVA F-values~\cite{Fisher92a}, reducing the dimensionality to the 35 most informative features per resolution.}

\review{Finally, we employ a CNN-based autoencoder adapted from~\cite{Wang19a}. The encoder uses stacked convolutional and pooling layers to capture local temporal dependencies in the load profiles, followed by a dense bottleneck layer producing compact embeddings. The decoder mirrors this structure with upsampling and deconvolutions to reconstruct the input sequence. These latent embeddings capture non-linear consumption patterns and enable a complementary perspective to handcrafted or ts-fresh features. However, similar to frequency coefficients in ts-fresh, autoencoder embeddings lack semantic interpretability and are therefore excluded from the interpretability analysis.}

\subsection{Classification}\label{subsection_classification}
Supervised machine learning techniques are employed to classify household-specific socio-demographic characteristics based on features extracted from weekly electricity load profiles. The classification task is formulated as a binary problem, where each model is trained to distinguish between two classes (positive and negative) corresponding to the presence or absence of a given characteristic.

\review{A key challenge when working with real-world data collected from field studies, such as the dataset described in Section~\ref{subsection_dataset}, lies in the imbalanced distribution of class labels for several characteristics. As shown in Table~\ref{tabel_dataset_description}, some characteristics are significantly underrepresented. For instance, only 256 households report owning a sauna, while 952 do not. This class imbalance can lead to biased models that favor the majority class, thereby degrading predictive performance, particularly for the minority class. Previous studies have demonstrated that such imbalances can adversely affect classification accuracy and generalization~\cite{Beckel13a,Beckel14a}.}

To mitigate this issue, we apply random undersampling during the training process. Undersampling has been shown to be an effective strategy for addressing class imbalance by equalizing the number of positive and negative samples in the training data~\cite{Japkowicz00a,He09a}. Specifically, we randomly remove samples from the majority class such that both classes contain the same number of instances, determined by the size of the minority class. Importantly, this resampling is applied only to the training and validation sets to preserve the original distribution in the test set for a realistic evaluation.

\review{For the classification task, we consider a diverse set of well-established binary classifiers, selected based on two main criteria: (i) comparability with existing work—particularly the studies by Beckel et al.~\cite{Beckel13a,Beckel14a} and Wang et al.~\cite{Wang19a}, and (ii) interpretability of learned feature importances, which is critical for privacy analysis. The latter motivates the inclusion of tree-based models such as XGBoost, which support post-hoc interpretability using SHAP values.}

\review{The final classifier set used in this study includes:
(i) Support Vector Machines (SVM): margin-based classifiers effective in high-dimensional spaces;
(ii) $k$-Nearest Neighbors (kNN): a non-parametric method that classifies based on feature space similarity;
(iii) AdaBoost: an ensemble method that combines weak learners to improve predictive performance;
(iv) Multi-Layer Perceptrons (MLP): simple feedforward neural networks capable of capturing non-linear decision boundaries;
(v) Linear Discriminant Analysis (LDA): a generative linear classifier that models class-conditional distributions; and
(vi) Extreme Gradient Boosting (XGBoost): a regularized boosting algorithm that additionally provides interpretable feature importance measures.}

\subsection{Evaluation Measures}\label{subsection_evaulation}
In the domain of supervised machine learning, the accuracy, defined as the ratio of the number of correct classifications to the total number of samples, is a commonly used metric for evaluating classifier performance~\cite{Sokolova09a}. 
The accuracy can be calculated as follows: 

\begin{equation}
    \text{ACC} \ =\ \ \frac{TP\ +\ TN}{TP\ +\ TN\ +\ FP\ +\ FN}.
\end{equation}

\textit{TP}, \textit{FN}, \textit{FP}, and \textit{TN} represent the number of samples that are correctly predicted as positive, incorrectly predicted as negative, incorrectly predicted as positive, and correctly predicted as negative. Precision and recall are calculated as follows:

\begin{gather}
    \text{$Precision$} \ =\ \ \frac{TP\ }{TP\ +\ FP\ }, 
    \ \text{$Recall$} \ =\ \ \frac{TP\ }{TP\ +\ FN}.
\end{gather}

Thus, the $F_1$ score is defined as: 
\begin{equation}
    \text{$F_1$} \ =\ \ 2 \cdot \frac{Precision\ \cdot \ Recall}{Precision\ +\ Recall}.
\end{equation}

Accuracy and $F_1$ score are commonly used statistics, which indicate respectively the proportion of true positives and true negatives relative to all elements and the mean of precision and recall. 
Both of these metrics are widely employed to asses binary classification models, but they tend to yield overly optimistic results, especially when dealing with datasets that exhibit a positive class imbalance~\cite{Chicco20a}. 

For a more informative evaluation, especially when dealing with an imbalanced dataset as mentioned in Section~\ref{subsection_classification} and shown in Table~\ref{tabel_dataset_description}, the MCC (Matthews Correlation Coefficient), also known as the phi coefficient, is computed. 
The coefficient takes into account true and false positives and negatives, making it a balanced measure suitable for the evaluation of imbalanced class sizes. 
MCC values range from -1 to +1, where +1 indicates a perfect classifier, 0 represents random predictions, and -1 signifies complete disagreement between the classifier's predictions and the actual labels~\cite{Matthews75a}. 

In the context of binary classification, the MCC is computed as follows:  
\begin{equation}
    \text{MCC} \ =\ \ \frac{TP\cdot TN-FP\cdot FN\ }{\sqrt{( TP+FP)( TP+FN)( TN+FP)( TN+FN)}}.
\end{equation}

\subsection{SHAP Values for Feature Importance}\label{subsection_feature_importance}

\review{To investigate which load-profile features contribute most to the prediction of socio-demographic characteristics, we employ SHAP (SHapley Additive exPlanations) values~\cite{Lundberg17a}, a method grounded in cooperative game theory. SHAP values allocate the output of a classifier to individual input features by quantifying their marginal contribution relative to a baseline expectation.}

\review{Given a classifier $f^{\dt}$ trained to predict a socio-demographic attribute $y$ from a weekly consumption vector $c^{\tilde{w},\dt} \in \R^n$, the SHAP value $\phi_j$ for feature $j \in F$ is formally defined as:}

\begin{equation}
\phi_j = \sum_{S \subseteq F \setminus \{j\}} 
\frac{|S|! \, (|F|-|S|-1)!}{|F|!} 
\Big( f^{\dt}(S \cup \{j\}) - f^{\dt}(S) \Big),
\end{equation}

\review{where $F$ denotes the set of all features and $S$ a subset of $F$ excluding $j$. This formulation ensures a fair distribution of predictive contributions, analogous to Shapley values in cooperative game theory.}

\review{SHAP values can be positive or negative. A positive $\phi_j$ indicates that feature $j$ increases the likelihood of the positive class (e.g., \textit{household has a swimming pool}), while a negative $\phi_j$ reflects a contribution towards the negative class. The absolute magnitude $|\phi_j|$ expresses the strength of the contribution.}

\review{In the context of our threat model, SHAP values provide interpretability at two levels: (i) global feature importance, by ranking features according to their mean absolute SHAP values across households and weeks, and (ii) local explanations, by quantifying the contribution of each feature to the prediction for a specific household-week $c^{\tilde{w},\dt}$.}

\section{Results}
\label{sec:Results}
\review{This section presents the experimental results obtained from analyzing the impact of time granularity on the prediction of household-specific socio-demographic characteristics. To comprehensively assess this relationship, we evaluate multiple dimensions of the prediction task. First, we investigate how varying temporal resolutions of weekly load profiles influence classification performance. Next, we compare different feature extraction strategies, contrasting interpretable handcrafted features with automatically generated ts-fresh features and representation learning via autoencoders. Building on this, we analyze the performance of several supervised learning algorithms to identify which classifiers are most effective in this context. To deepen our understanding, we further examine feature importance and interpretability, highlighting which load-profile characteristics drive predictive power across time resolutions.}

\subsection{Influence of Time Granularity}
\begin{figure*}[ht]
  \centering
  \includegraphics[width=\linewidth]{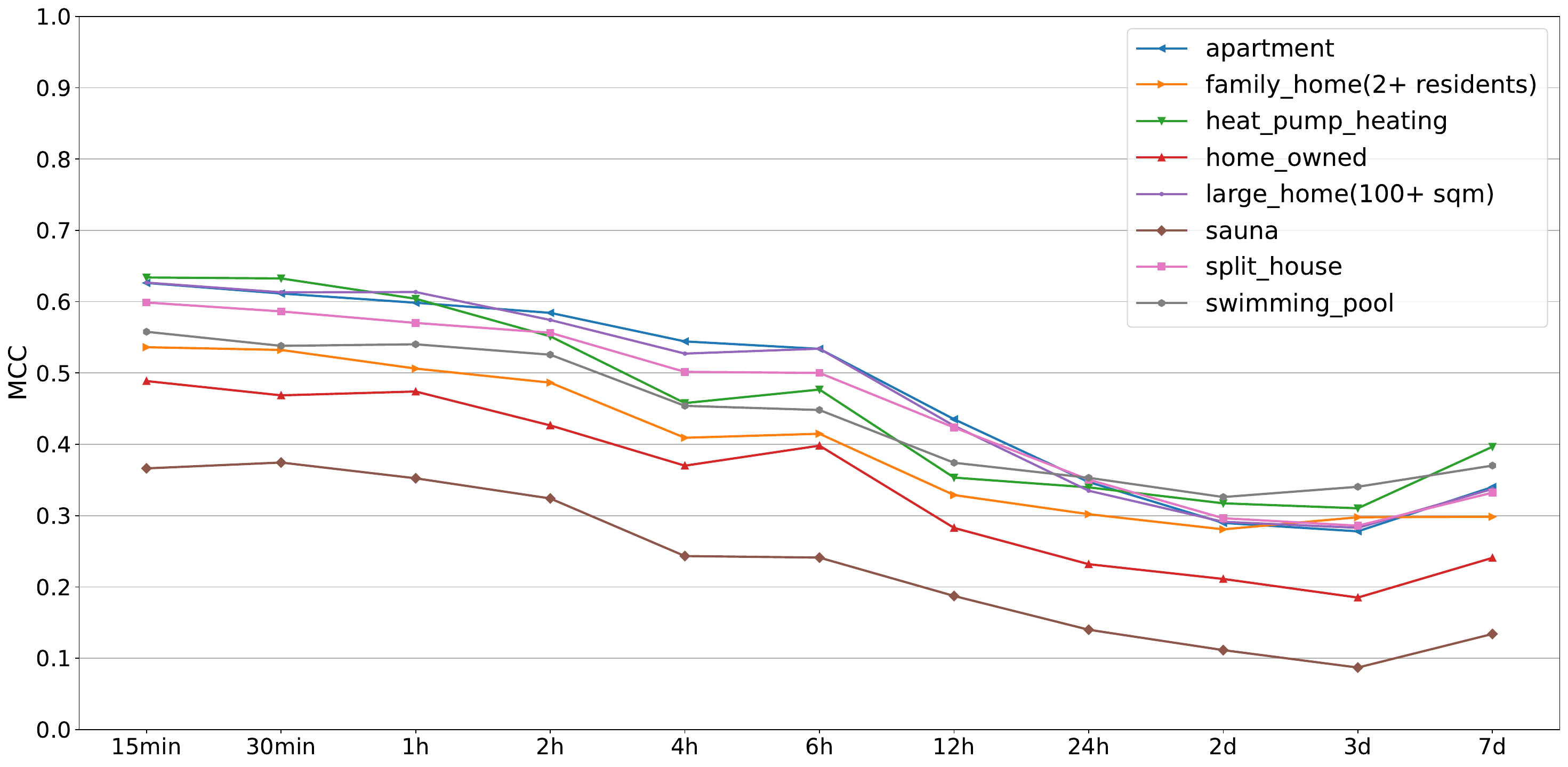}
  \caption{Matthews correlation coefficient (MCC) of the \texttt{XGBoost} classifier using handcrafted features across all temporal granularities, as presented similarly in~\cite{Radovanovic25a}. Colors indicate the socio-demographic characteristics shown in the legend}
  \label{figure_results_MCC_score_for_xgboost}
\end{figure*}

\begin{figure*}[h!]
  \centering
  \includegraphics[width=\linewidth]{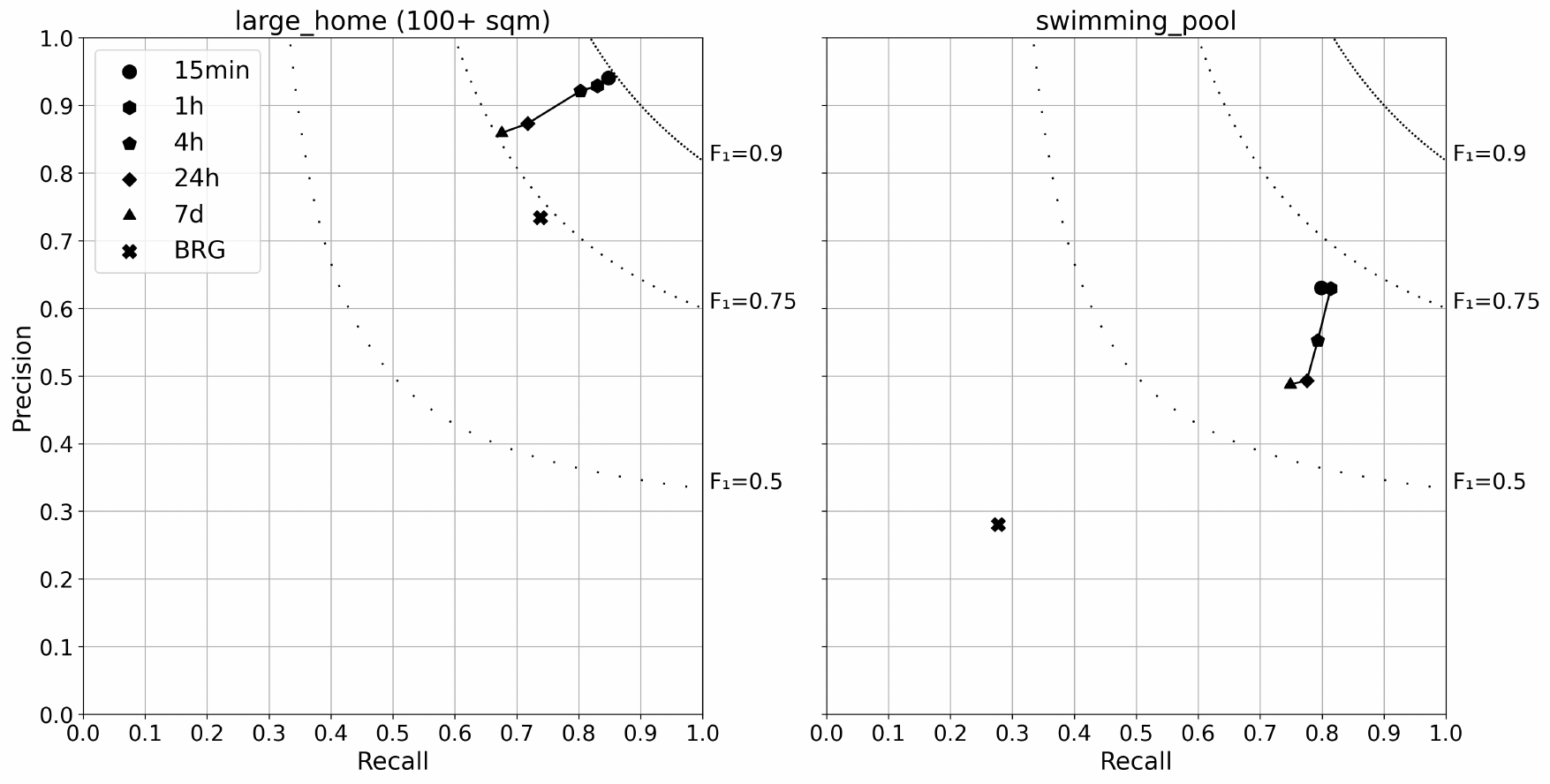}
  \caption{Prediction performance of the \texttt{XGBoost} classifier shown as precision–recall plots for the socio-demographic attributes \textit{large\_home} and \textit{swimming\_pool}, as previously presented in~\cite{Radovanovic25a}. Symbols denote different time granularities, while the cross (×) marks the baseline of biased random guessing.}
  \label{figure_results_precision_recall_large_home}
\end{figure*}

\review{For the analysis of the time granularity influence on the prediction performance of household-specific socio-demographic characteristics, we employ two complementary evaluation perspectives. First, the Matthews Correlation Coefficient (MCC) is used as a robust single-value metric, capturing both sensitivity and specificity. Figure~\ref{figure_results_MCC_score_for_xgboost} illustrates the MCC-based prediction performance of the \texttt{XGBoost} classifier across all time granularities. Second, a precision–recall analysis highlights the trade-off between precision and recall for individual characteristics, exemplified in Figure~\ref{figure_results_precision_recall_large_home}.}

\review{According to~\cite{Chicco21a}, MCC values can be interpreted analogously to correlation strength: values above 0.7 indicate a strong relationship, values between 0.7 and 0.4 represent moderate correlation, values between 0.4 and 0.2 indicate a weak relationship, and values below 0.2 are comparable to random guessing. For example, the characteristic \textit{family\_home} achieves an MCC of 0.57 at 15-minute resolution, which places it in the moderate range. The performance decreases slightly to 0.54 at one hour and drops further into the weak correlation range at coarser resolutions.}

\review{Between two and 24 hours, most socio-demographic characteristics exhibit significant declines, although individual patterns vary. For instance, \textit{swimming\_\\pool} shows a strong drop between six and twelve hours followed by a slight recovery, whereas \textit{sauna} increases marginally in the same range before dropping at 24 hours. Beyond the daily resolution, prediction performance stabilizes with no major changes across characteristics.}

\review{A similar behavior is also observed for the automatically generated \texttt{ts-fresh} features, where prediction performance declines consistently with coarser resolutions (see Section~\ref{figure_results_MCC_score_for_xgboost_ts_fresh}). A comparable trend is found across all classifiers. For example, the MCC for \textit{heat\_pump\_heating}, \textit{apartment}, and \textit{large\_home} decreases from about 0.65 at 15 minutes to roughly 0.36 with \texttt{XGBoost}, compared to a range of 0.58 to 0.38–0.41 for \texttt{AdaBoost}, and a similar decline for \texttt{SVM} from about 0.6 to 0.37, as discussed in Section~\ref{subsec:classifier_performance}.}

\review{Figure~\ref{figure_results_precision_recall_large_home} complements the MCC analysis by providing a precision–recall evaluation for the socio-demographic characteristics \textit{large\_home} (left) and \textit{swimming\_pool} (right) using the \texttt{XGBoost} classifier. These labels were deliberately chosen because they are frequently examined in related work (see Section~\ref{sec:Discussion}) and represent two privacy-relevant attributes\cite{Beckel13a,Wang19a,Wang20b}: \textit{large\_home} reflects a more static consumption behavior, while \textit{swimming\_pool} captures a dynamic usage pattern. The x-axis denotes recall (true positive rate), while the y-axis represents precision (positive predictive value). Black markers trace the evolution of performance across time granularities, with the cross (X) marking the baseline of biased random guessing, which increases with greater class imbalance. In line with the MCC results, both precision and recall decline as the time resolution becomes coarser. The effect is less pronounced for \textit{large\_home}, which shows more stable consumption patterns, whereas \textit{swimming\_pool} exhibits stronger variability. Despite these declines, the classifier consistently outperforms random guessing.}

\review{For enhanced inspection, Figure~\ref{figure_results_precision_recall_large_home_zoomed} in the Appendix provides a zoomed-in version of all resolutions. Since the baseline of biased random guessing lies far outside the region of interest, it is omitted from this detailed view to allow clearer comparison of the classifier’s performance across time granularities.}

\subsection{Comparison of Feature Extraction Methods}
\review{To assess the impact of different feature representations, we compare the performance of three feature extraction approaches across all selected socio-demographic characteristics: (i) handcrafted features (blue), (ii) automatically generated features using the \texttt{ts-fresh} library (orange), and (iii) latent embeddings derived from a CNN-based autoencoder (green). For this analysis, the \texttt{XGBoost} classifier was applied consistently to ensure comparability between methods.}

\begin{figure*}[ht]
  \centering
  \includegraphics[width=\linewidth]{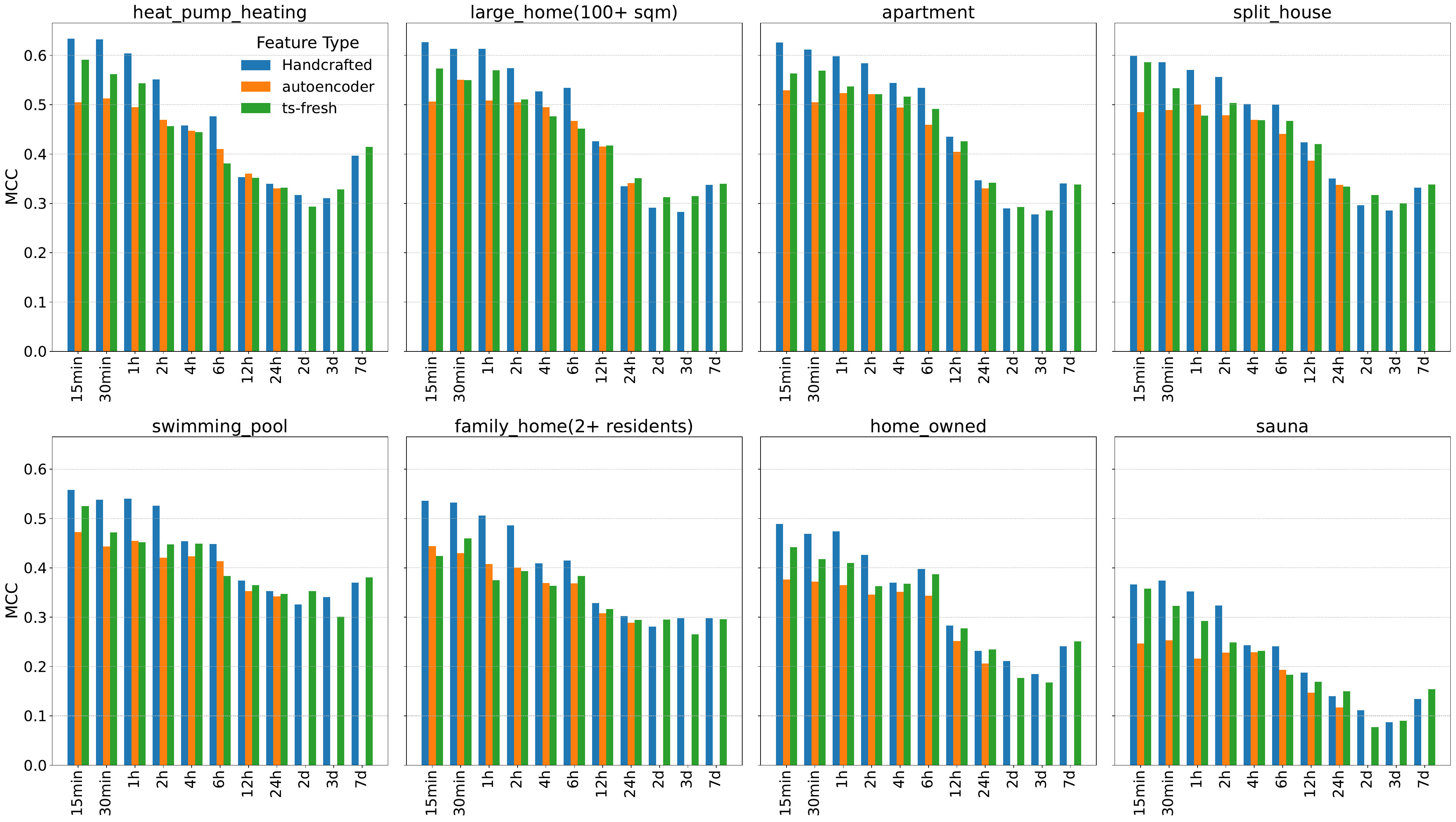}
  \caption{\review{MCC for the classification result of the \texttt{XGBoost} classifier for all time granularities. The colors represent the time granularity listed in the legend.}}
  \label{figure_results_feature_extraction_methods}
\end{figure*}

\review{Figure~\ref{figure_results_feature_extraction_methods} illustrates the results in the form of subplots, one for each socio-demographic characteristic. Within each subplot, the prediction performance is shown as MCC values across all considered time granularities. While handcrafted and \texttt{ts-fresh} features could be computed at all resolutions, the autoencoder approach was restricted to granularities up to one day, since coarser resolutions (\mbox{two days, three days, and seven days}) could not be processed with the convolutional structure of the model.}

\review{The results in Figure~\ref{figure_results_feature_extraction_methods} highlight systematic differences between feature extraction strategies. Across most socio-demographic characteristics, handcrafted and \texttt{ts-fresh} features yield comparable performance, with handcrafted showing slight advantages at finer resolutions (15–60 minutes) for attributes with higher temporal variability, such as \textit{swimming\_pool} and \textit{sauna}. \texttt{ts-fresh}, in contrast, often maintain stable performance across a broader range of granularities, particularly for more static attributes such as \textit{apartment} or \textit{large\_home}. Autoencoder embeddings achieve competitive results at high temporal resolutions (up to 24 hours) but cannot be applied to coarser resolutions, limiting their applicability. Overall, the comparison demonstrates that while automated feature generation and representation learning offer strong predictive potential, especially at fine resolutions, handcrafted features remain robust and interpretable across the entire spectrum of granularities. This underlines the importance of aligning feature extraction strategies with the temporal characteristics of the prediction task.}

\subsection{Classifier Performance Analysis}\label{subsec:classifier_performance}
\begin{figure*}[ht]
  \centering
  \includegraphics[width=\linewidth]{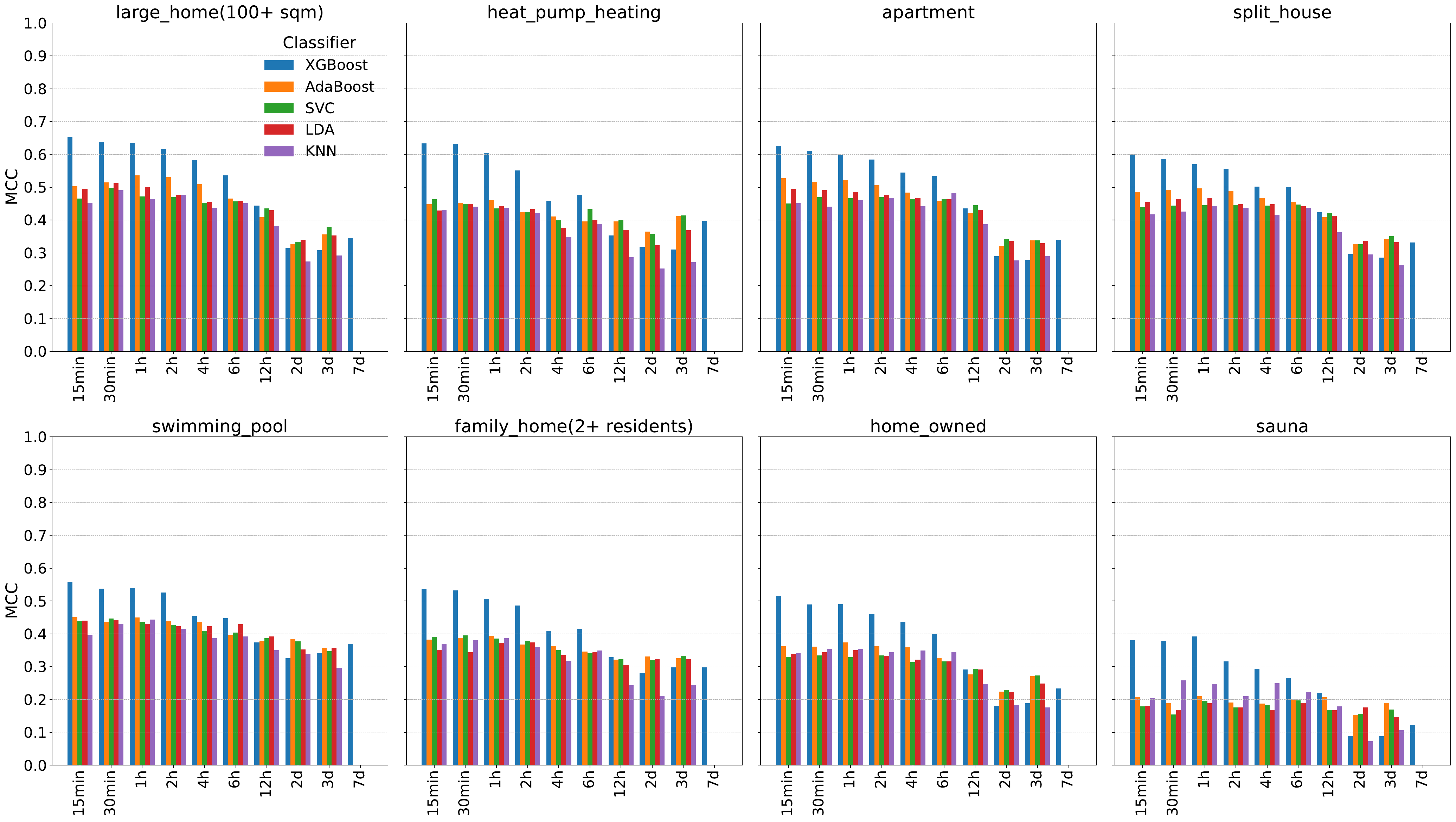}
  \caption{\review{MCC for the classification result of the \texttt{XGBoost} classifier for all time granularities. The colors represent the time granularity listed in the legend.}}
  \label{figure_results_classifier_comparison}
\end{figure*}

\review{After analyzing the effect of temporal resolution and feature representation, we now examine how the choice of classifier influences prediction performance. To this end, we evaluate a set of commonly used supervised learning algorithms, including \texttt{XGBoost}, \texttt{AdaBoost}, \texttt{Support Vector Classifier (SVC)}, \texttt{Linear Dis \\ criminant Analysis (LDA)}, and \texttt{k-Nearest Neighbors (KNN)}. Figure~\ref{figure_results_classifier_comparison} depicts the classification results for all socio-demographic characteristics when using handcrafted features as input across the full range of temporal granularities. This analysis provides a systematic comparison of classifier robustness and highlights their relative effectiveness in the context of socio-demographic inference from coarse-grained load profiles.}

\review{The results in Figure~\ref{figure_results_classifier_comparison} highlight clear differences in classifier robustness when applied to handcrafted features across all temporal granularities. Overall, \texttt{XGBoost} consistently achieves the highest predictive performance, particularly at finer granularities (15–60 minutes), where MCC values frequently reach the strong range. \texttt{AdaBoost} and \texttt{SVC} generally follow as second-tier methods, showing comparable trends but at lower performance levels. \texttt{LDA} demonstrates moderate results, performing reasonably well at fine resolutions but declining more sharply as granularity increases. In contrast, \texttt{KNN} exhibits the weakest performance across nearly all socio-demographic characteristics. While all classifiers show declining accuracy as profiles are aggregated into longer intervals, the relative ranking between models remains stable, underlining \texttt{XGBoost}’s superior robustness for socio-demographic inference from coarse-grained load profiles. It should be noted, however, that for some classifiers the coarsest resolution of seven days could not be computed, reflecting limitations in model applicability at extreme aggregation levels.}

\subsection{Feature Importance and Interpretability}
\review{To complement the analysis of classifier performance, we investigate which handcrafted features of the load profiles most strongly influence the prediction of socio-demographic characteristics. We restrict our analysis to handcrafted features because their meaning is interpretable in terms of consumption behavior (e.g., minimum load, evening consumption). In contrast, automatically generated \texttt{ts-fresh} features, such as FFT or CWT coefficients, are more difficult to interpret, and autoencoder embeddings do not permit direct feature-level attribution.}

\begin{figure*}[ht]
  \centering
  \includegraphics[width=\linewidth]{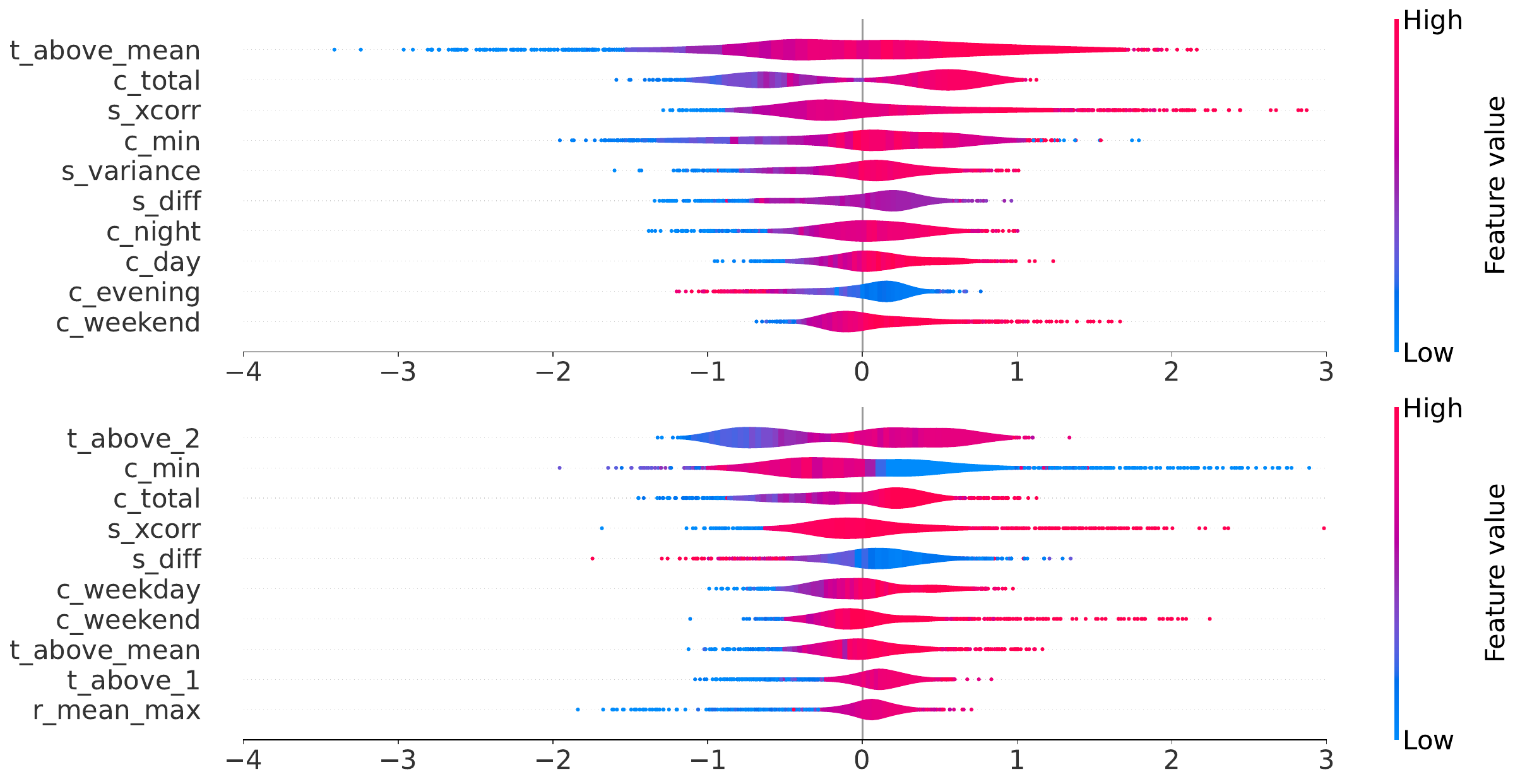}
  \caption{\review{Violin plots of SHAP feature importance for the socio-demographic characteristic \textit{swimming\_pool} using handcrafted features. The top plot shows results at 15-minute granularity, the bottom plot at 4-hour granularity. The y-axis lists the ten most important features, while the x-axis depicts SHAP values indicating the impact on the model output.}}
  \label{fig:FeatureImportance_SwimmingPool}
\end{figure*}

\review{Figure~\ref{fig:FeatureImportance_SwimmingPool} presents violin plots of SHAP values for the ten most important handcrafted features predicting the presence of a \textit{swimming\_pool}, shown at two temporal resolutions: 15 minutes (top) and 4 hours (bottom). In these plots, the y-axis lists the ten features with the highest contributions to the model, while the x-axis depicts their SHAP values, quantifying the marginal impact of each feature on the model output. Positive values increase the likelihood of predicting swimming pool ownership, whereas negative values support the opposite. The violin shapes indicate the distribution of contributions across households, and the color gradient encodes the feature values (red = high, blue = low).}

\review{At both 15-minute and 4-hour resolutions, a largely overlapping set of features emerges as important for predicting swimming pool ownership, yet their relative contributions and the structure of their SHAP distributions differ. At 15 minutes, features such as \textit{t\_above\_mean}, \textit{c\_total}, and \textit{s\_diff} dominate. The feature \textit{t\_above\_mean} depicts high values concentrated between –1 and 1, while \textit{c\_total} is characterized by two distinct violin shapes around –1 and 1, reflecting the strong contrast in total weekly consumption between households with and without pools. In addition, \textit{s\_diff} indicates high mid-range values, highlighting the importance of signal variability at fine resolutions.}

\review{At 4 hours, similar features remain relevant, but their patterns shift due to temporal aggregation. For instance, \textit{t\_above\_2} (time above 2 kWh) exhibits two violin distributions around –1 and 1, centered closer to zero, indicating a weaker separation compared to the finer granularity. The feature \textit{c\_min} contributes mainly with lower positive SHAP values, suggesting that minimum consumption levels provide only limited discriminative power at this resolution. A comparable attenuation is observed for \textit{s\_diff}, where the spread of values narrows substantially.}

\review{Taken together, these results show that while the same set of consumption indicators is consistently identified as important across granularities, temporal aggregation dampens their discriminative strength. Fine-grained features highlight sharp differences in consumption behavior (e.g., evening and night pump operation), whereas at coarser resolutions, the predictive signal becomes weaker and less distinct.}

\review{A similar analysis for the socio-demographic characteristic \textit{large\_home} is depicted in Figure~\ref{fig:FeatureImportance_Large_Home}. At 15-minute granularity, features such as \textit{c\_min} and \textit{c\_evening} emerge as dominant, reflecting the consistently higher baseline and elevated evening consumption of larger households. At 4-hour resolution, the feature ranking shifts: \textit{c\_min} remains important but is complemented by aggregate indicators such as \textit{c\_max} and temporal threshold features (e.g., \textit{t\_above\_1}, \textit{t\_above\_0.5}, \textit{t\_above\_2} and \textit{t\_above\_mean}). The narrowing of SHAP distributions indicates reduced variability in feature contributions, similar to the swimming pool case, but here the predictive signal remains more stable, as dwelling size is a comparatively static attribute.}

\begin{figure*}[ht]
    \centering
    \includegraphics[width=\linewidth]{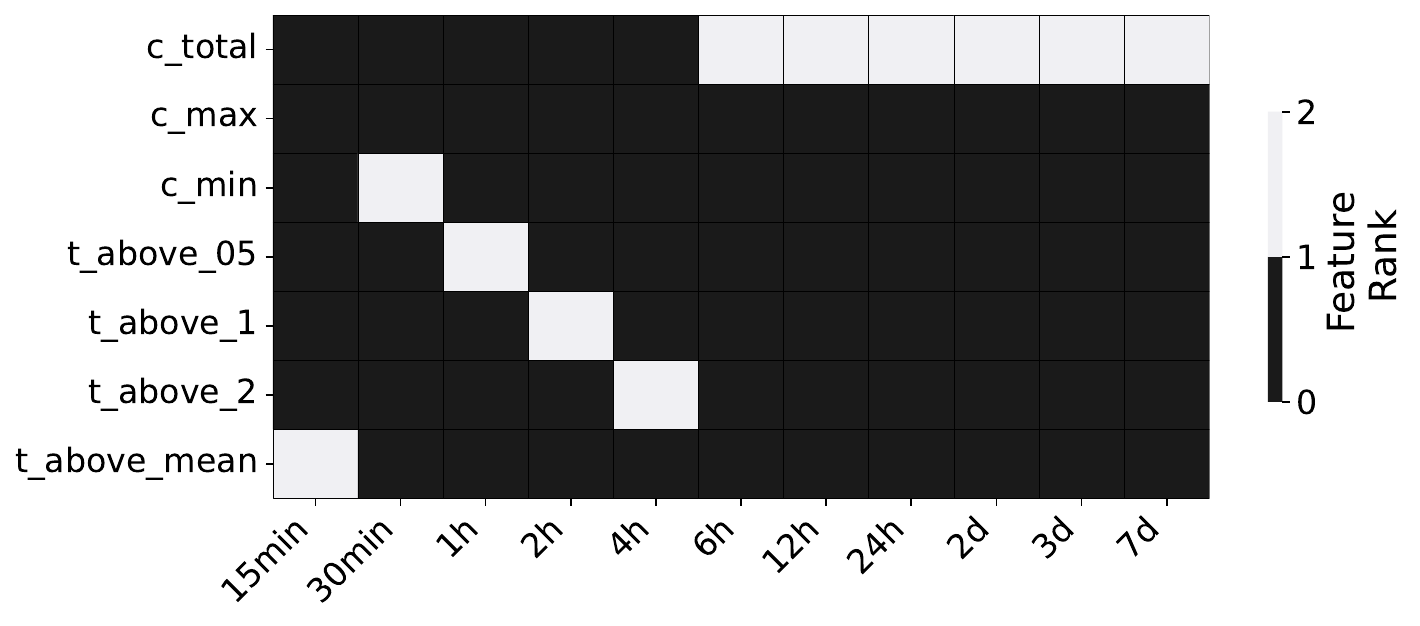}
    \caption{\review{Heatmap of the most influential handcrafted feature for predicting the socio-demographic attribute \textit{swimming\_pool} across all temporal granularities. Each column corresponds to one granularity, and the lightest cell in a column indicates the feature ranked highest in terms of SHAP importance. Dark cells denote that a feature was not the most dominant at the respective resolution.}}
  \label{fig:swimming_pool_ranks_heatmap}
\end{figure*}

\review{To extend this comparison across all temporal resolutions, Figures~\ref{fig:swimming_pool_ranks_heatmap} and~\ref{fig:large_home_ranks_heatmap} present heatmaps of the most influential handcrafted feature per granularity. For \textit{swimming\_pool}, dominance shifts markedly with coarsening resolution, from dynamic intra-day consumption features at fine granularities to total or maximum consumption aggregates at daily and weekly levels. In contrast, for \textit{large\_home}, the most influential feature is consistently \textit{c\_min} across most resolutions, with occasional shifts to other aggregates (e.g., \textit{t\_above\_1} at 6 hours, \textit{c\_total} at 12 hours, \textit{c\_weekday} at 24 hours). This comparison illustrates a key distinction: dynamic socio-demographic attributes such as swimming pool ownership are highly sensitive to temporal resolution, whereas static attributes like dwelling size can be reliably inferred even from coarser profiles.}

\section{Discussion and Comparison to Related Work}
\label{sec:Discussion}
If a decision has to be made concerning the granularity of collected load data, privacy implications of load-profile time granularities for a specific households, our findings suggest that use cases which require data minimization or maximum privacy may use one-hour data without loss of classification performance for those seeking more utility in their load profiles, a time granularity of one hour strikes a balance between data utility and privacy preservation. 
Conversely, for use cases which require maximum granularity (15 minutes), e.g., for legal or regulatory reasons, no better prediction performance is achieved by using higher granularity those prioritizing heightened privacy, a time granularity of one week offers significant privacy preservation. 

Best results regarding prediction performance are exhibited for the household characteristics \textit{heat\_pump\_heating}, \textit{apartment} and \textit{large\_home} with values between $0.65$ and $0.6$, when relying on weekly snippets with 15-minute time granularity. 
While varying MCC values are reported for the eight examined household characteristics, the consistent decline in the prediction performance across all characteristics indicates that the type of characteristic being predicted is not the most significant factor in the trade-off between data utility and privacy preservation. Instead, it suggests that the increase in time granularity may play a more significant role in the change of the prediction performance.

A first limitation lies in the ground truth: questionnaire-based socio-demo- graphic information may be inaccurate or incomplete, and week-specific appliance usage (e.g., swimming pools, saunas) is not available. We therefore assigned constant yearly labels to all weeks, which may have introduced mislabels, especially for dynamic attributes. Nevertheless, classification results across all attributes remained substantially above random guessing. A second limitation is the feature design: the handcrafted features originate from a 30-minute setting~\cite{Beckel14a} and were applied to coarser resolutions, even up to seven days. Defining new features tailored to such long intervals remains a direction for future research.

The methodological distinction from~\cite{Beckel14a} is critical. While their evaluation trained and tested on the same fixed week of the year, our design trains on year-round data and predicts arbitrary weeks, requiring models to generalize across seasonal and behavioral variations. Despite this greater complexity, our approach achieves higher performance, e.g., up to 0.65 MCC compared to <0.4 MCC at 30 minutes for \textit{family\_home} (0.34), \textit{large\_home} (0.18), and \textit{housetype} (0.20). To ensure comparability, we also applied the handcrafted feature extraction method of~\cite{Beckel14a} and evaluated it across multiple classifiers. As shown in Figure~\ref{figure_results_classifier_comparison}, \texttt{XGBoost} consistently outperforms all classifiers used in~\cite{Beckel14a} for every socio-demographic characteristic.

% As already stated in Section~\ref{sec:Related_Work}, it was not clear, whether our methodology that uses training data of one year and predicts a single, arbitrary week of the year is better than the one from~\cite{Beckel14a} where training and testing data are from the same, single week of the year. It turns out that our approach leads to better performance values: the figures in~\cite{Beckel14a} show MCC values <0.4 for the labels familyhome (0.34), largehome (0.18), and housetype (0.2), classified with the SVM. Our approach leads to values up to 0.65 at the same time granularity of 30 min which are consistently higher for those labels.
% While the definition of the labels are not exactly the same it should be noted that (i) we tried to mimic the labels from~\cite{Beckel14a} as good as possible to enable a fair comparison (for example we did not have information about children in our dataset) and (ii) the choice of the thresholds did not include any kind of optimization with respect to classification performance.

The impact of time granularity on socio-demographic inference differs markedly from appliance detection, as shown in~\cite{Eibl15a}. In their study, with granularities as fine as 3 seconds, privacy was largely preserved already at 15 minutes, except for lighting usage, and privacy arose mainly from reduced recall with stable precision. By contrast, in our setting both recall and precision decline with coarser granularity (Figure~\ref{figure_results_precision_recall_large_home}, right plot). A comparison with~\cite{Ferner19a} and~\cite{Burkhart18a}, who predicted pool ownership using full-year time series, further highlights the competitiveness of our approach: despite relying on only a single week of data and standard handcrafted features from~\cite{Beckel13a}, we achieve an accuracy of 0.81 and precision of 0.63 at 15-minute resolution, close to their reported 0.93 accuracy and 0.67 precision with SVM classifiers.

Beyond performance trends, our study provides methodological insights regarding feature extraction, classifier choice, and interpretability. Feature extraction analysis showed that handcrafted and ts-fresh features yield comparable accuracy across granularities, while autoencoder embeddings—though less interpretable—remain competitive up to daily resolution. This confirms that interpretable feature sets can rival more complex representations without sacrificing predictive power. Furthermore, feature importance analysis revealed systematic differences between static and dynamic attributes: for static characteristics such as \textit{large\_home}, baseline load indicators like \textit{c\_min} consistently dominated across resolutions, reflecting persistent background consumption. In contrast, for dynamic attributes such as \textit{swimming\_pool}, short-term features (e.g., \textit{c\_night}, \textit{s\_diff}) were decisive at fine resolutions, whereas at coarser scales aggregate indicators (e.g., \textit{c\_total}) took precedence. This demonstrates that static household properties can be robustly inferred even at coarse granularity, while dynamic attributes are substantially more sensitive to temporal aggregation.

% The influence of time granularity on socio-demographic features is very different from the one of appliance detection as a comparison with~\cite{Eibl15a} shows. There the time granularity starts at 3s, and except for usage of lights, privacy is already very well preserved for our \textit{coarsest} granularity of 15 minutes. Privacy is also achieved differently then here: while their privacy is comes mainly from a decreasing recall with rather stable precision, here both, recall and precision, decrease with coarser time granularity (Figure~\ref{figure_results_precision_recall_large_home}).

% The prediction performance of our methodology can also be compared to~\cite{Ferner19a} and~\cite{Burkhart18a} where the existence of a pool was predicted. Although both~\cite{Ferner19a} and~\cite{Burkhart18a} assume that training and test data are timeseries of a whole year, our approach is quite competitive despite the usage of just a single week of the year as input and the fact that we only took the standard features from~\cite{Beckel13a}: while for \texttt{SVM} Gaussian and handcrafted features Ferner et al. achieved no pool classification 0.93 accuracy and 0.67 precision, our overall precision is 0.63 and the overall accuracy 0.81 using the same 15 minute granularity as in~\cite{Ferner19a} and ~\cite{Burkhart18a} (Figure \ref{figure_results_precision_recall_large_home}, right plot).

\section{Conclusion and Outlook}
\label{sec:Conclusion}
We introduce a novel evaluation methodology for predicting household-specific socio-demographic characteristics from load profiles with varying time granularities, based on randomly selected weekly snippets across one year. This distinguishes our approach from prior work that either relies on a fixed week for both training and evaluation or uses entire yearly profiles. Despite this more complex prediction setting, our methodology yields superior performance for several socio-demographic attributes compared to earlier studies, even when applying the same handcrafted feature extraction method.

Our findings confirm that predictive performance generally declines as granularity becomes coarser, yet two plateaus emerge. First, one-hour resolution achieves results comparable to 15 minutes, suggesting that regulatory requirements for 15-minute metering provide little additional predictive power over hourly aggregation. Second, beyond 24 hours, performance stabilizes up to 7 days, reflecting the limits of handcrafted features in capturing meaningful variability at very coarse resolutions. These plateaus imply that, depending on the application, finer resolutions do not necessarily yield significant utility gains and coarser resolutions can be chosen to enhance privacy in line with the principle of data minimization.

Beyond performance trends, our study provides methodological insights. Feature extraction analysis demonstrated that interpretable feature sets—both handcrafted and ts-fresh—remain competitive with learned embeddings, which perform well up to daily resolution but lack interpretability. Classifier comparisons showed that \texttt{XGBoost} consistently outperforms alternatives across all attributes and granularities, underlining its robustness in this problem space. Feature importance analysis further revealed a systematic distinction between static and dynamic socio-demographic characteristics: static attributes such as \textit{large\_home} are strongly driven by persistent baseline load (\textit{c\_min}), while dynamic attributes such as \textit{swimming\_pool} rely on fine-grained temporal features (\textit{c\_night}, \textit{s\_diff}) at high resolution and shift toward aggregate indicators (\textit{c\_total}) at coarser scales. This demonstrates that static household properties can be reliably inferred even at coarse resolutions, whereas dynamic attributes are more sensitive to temporal aggregation.

Finally, there is significant room for improvement concerning the correct matching of weekly load profiles to their associated socio-demographic household characteristics. 
We assume the characteristics to be constant for the whole year, even if the presence of some characteristics (e.g. the use of appliances such as \texttt{sauna} or \texttt{swimming\_pool}, or the number of residents present in a given week) may fluctuate over the course of a year, leading to incorrect training results for the classifier employed. Moreover, the handcrafted features used here were optimized for sub-daily resolutions and may fail to capture patterns beyond 24 hours. Future work should therefore design features tailored to coarse resolutions, incorporate datasets with week-level ground truth,

\begin{credits}
\subsubsection{\ackname} The financial support by the Federal State of Salzburg is gratefully acknowledged.
\end{credits}

\bibliographystyle{splncs04}
\bibliography{mybibliography}

\newpage
\appendix

\section{Implementation Details}
\review{All experiments are implemented in Python. Data handling, splitting, and classical models use \texttt{scikit-learn}; gradient-boosted trees use \texttt{XGBoost}; the convolutional autoencoder is implemented in \texttt{TensorFlow}/\texttt{Keras}; SHAP-based interpretability uses \texttt{shap}. The code is available at: \url{https://its-git.fh-salzburg.ac.at/fhs38238/time_resolution}.}

\paragraph{Data splitting and class balancing.}
For each label, we evaluate on unseen data via disjoint train/validation/test splits. We first hold out $10\%$ of all weekly snippets as a test set used only for final evaluation. From the remaining data, we subsample the training split to a balanced $50{:}50$ ratio of positives/negatives for the target label (random undersampling of the majority class). Of the balanced training split, $20\%$ is used as a validation set (for early stopping / model selection). This protocol ensures (i) strict separation of test data, (ii) consistent class balance during training, and (iii) availability of an unbiased validation signal.

\paragraph{Classifier configurations (constant across labels and granularities).}
We use fixed hyperparameters across all experiments to isolate the effect of time granularity and feature representation from classifier tuning.

\begin{itemize}
\item XGBoost (\texttt{XGBClassifier}): \texttt{max\_depth} $=5$, \texttt{n\_estimators} $=400$, \\\texttt{objective} \texttt{binary:logistic}, \texttt{importance\_type} \texttt{gain}, \texttt{eval\_metric} \\\texttt{['error','logloss']}. Early stopping is enabled via a callback, e.g.\\ \texttt{xgboost.callback.EarlyStopping(rounds=10, save\_best=True)}, monitored on the validation set passed through \texttt{eval\_set}.
\item AdaBoost: \texttt{AdaBoostClassifier} with \texttt{n\_estimators} $=200$ (default base estimator).
\item SVC: RBF kernel with probability estimates enabled (\texttt{probability=True}). Inputs are standardized by z-score (\texttt{StandardScaler} fitted on training data, applied to validation/test).
\item KNN: \texttt{KNeighborsClassifier} with $k=5$.
\item LDA: \texttt{LinearDiscriminantAnalysis} with default priors.
\end{itemize}

\paragraph{Convolutional autoencoder (feature learning).}
For resolutions up to and including 24,h, we learn latent embeddings with a lightweight CNN autoencoder (in the spirit of~\cite{Wang19a}). Inputs are standardized (z-score) per feature using training statistics. The encoder applies small receptive-field convolutions and downsampling; the decoder mirrors this with transposed convolutions to reconstruct the input. The concrete setup we use is:

\begin{itemize}
  \item \textbf{Hyperparameters:} latent dimension $=32$, L2 weight decay $=10^{-4}$, dropout rate $=0.5$.
  \item \textbf{Encoder:} \texttt{Conv2D(8,(2,3),relu)}, \texttt{Conv2D(16,(3,3),relu)}, \\\texttt{MaxPooling2D(pool\_size=(2,2), padding='same')}, \texttt{Dropout(0.5)}, \\ \texttt{Conv2D(16,(3,3),relu)}, \texttt{Flatten()}, \texttt{Dense(32, activation='linear')} (latent).
  \item \textbf{Decoder:} \texttt{Dense(4 $\times$ $\lceil T/2 \rceil$ $\times$ 16, relu)} $\rightarrow$ \\\texttt{Reshape} $\rightarrow$ \texttt{Conv2DTranspose(16,(3,3),strides=(2,2),relu)} \\ $\rightarrow$ \texttt{Conv2D(8,(3,3),relu)} $\rightarrow$ \texttt{Conv2D(1,(3,3),activation='linear')}.
\end{itemize}

We train the autoencoder with a reconstruction loss (MSE) and L2 regularization, using early stopping on validation loss to prevent overfitting. The learned 32-d latent vectors are then used as features for downstream classification. (For coarser-than-daily resolutions—2,d, 3,d, 7,d—the CNN autoencoder is not applied due to the incompatibility between very short temporal grids and the chosen convolution/pooling scheme.)

\section{Additional Results on Time Granularity}
\begin{figure*}[ht]
  \centering
  \includegraphics[width=\linewidth]{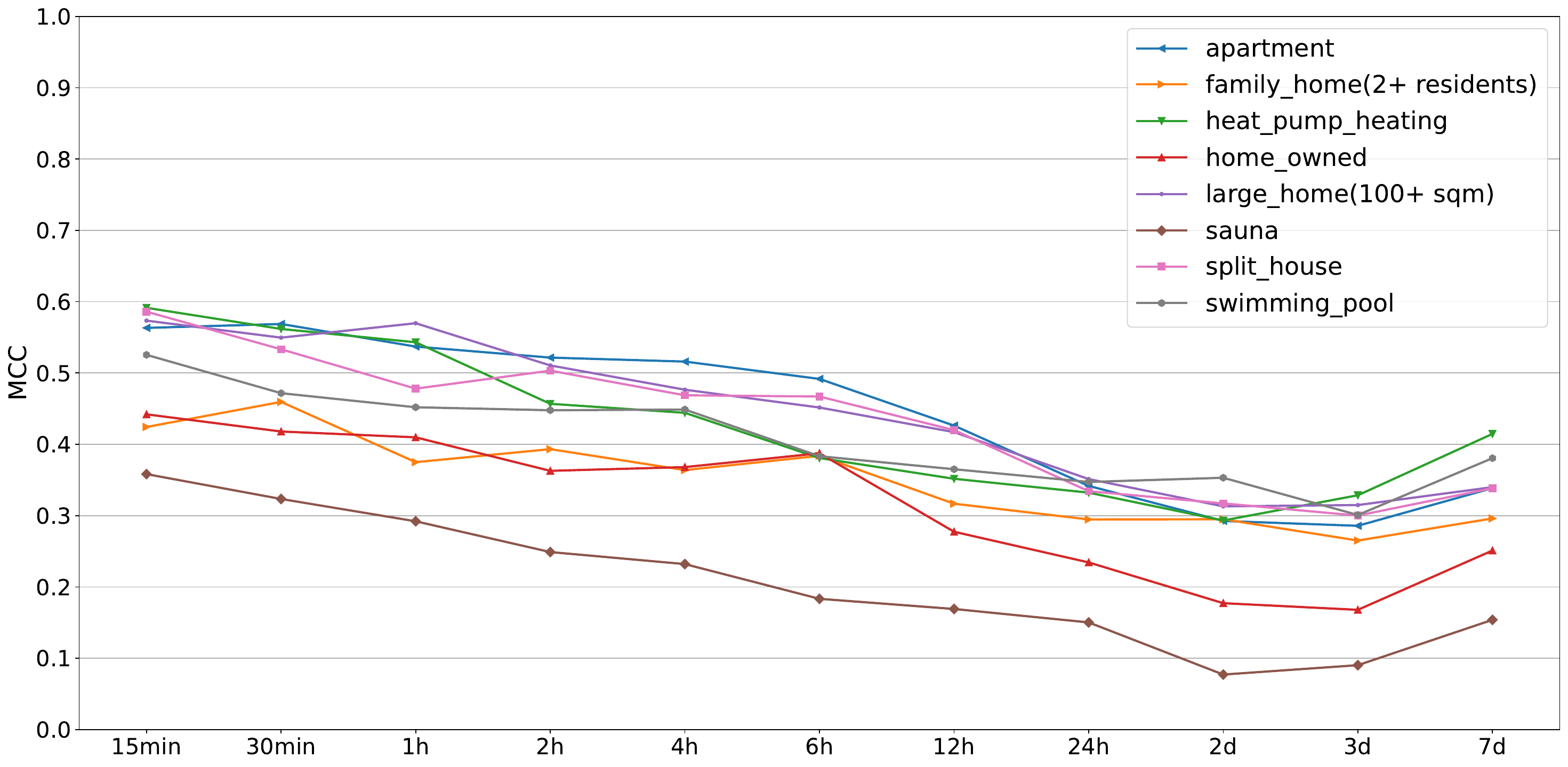}
  \caption{MCC results for \texttt{XGBoost} with \texttt{ts-fresh} features across all granularities.}
  \label{figure_results_MCC_score_for_xgboost_ts_fresh}
\end{figure*}

\begin{figure*}[h!]
  \centering
  \includegraphics[width=\linewidth]{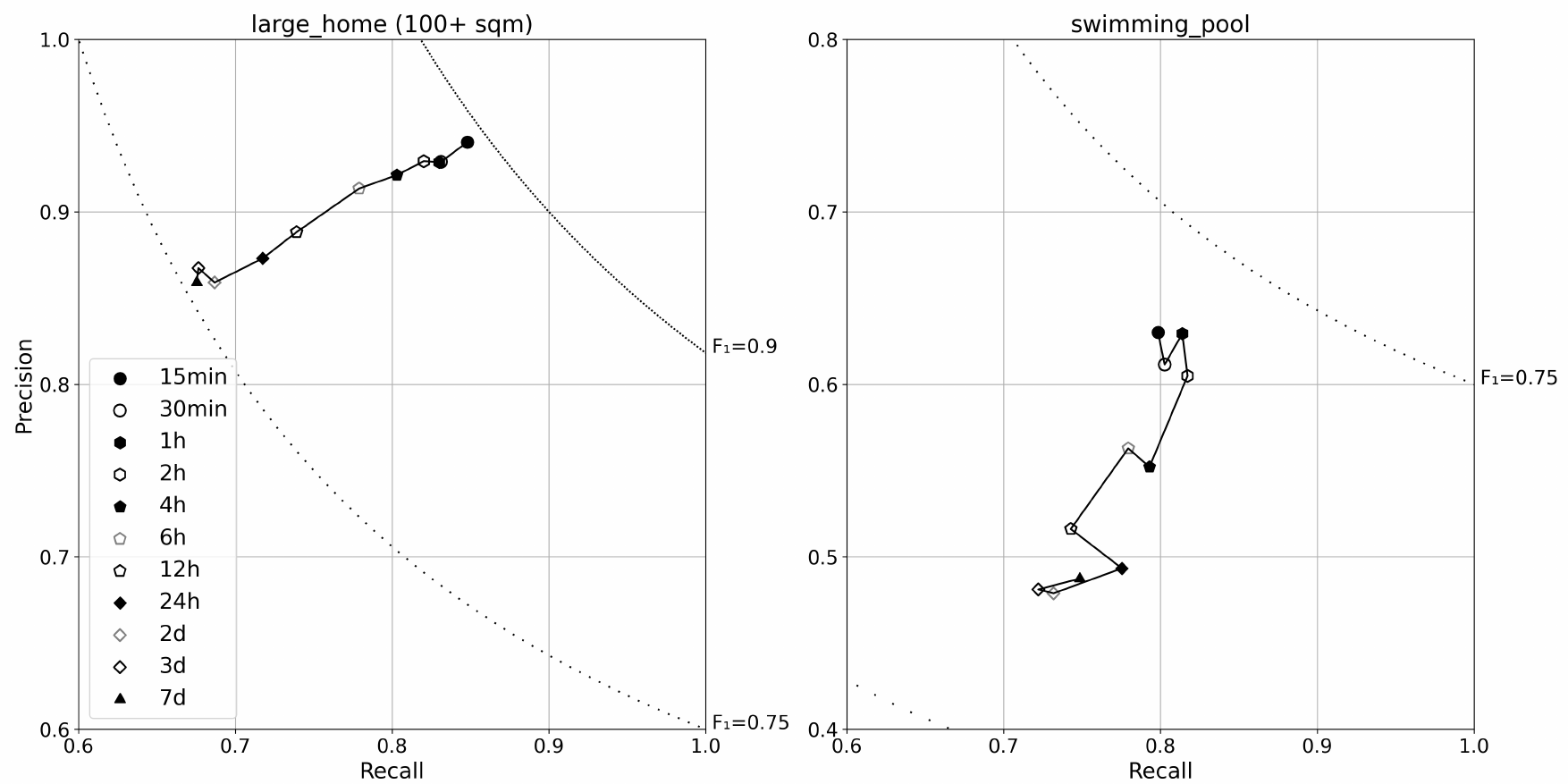}
  \caption{Precision–recall plots for the socio-demographic attributes \textit{large\_home} and \textit{swimming\_pool}, as presented similarly in~\cite{Radovanovic25a}}
\label{figure_results_precision_recall_large_home_zoomed}
\end{figure*}

Figure~\ref{figure_results_MCC_score_for_xgboost_ts_fresh} shows the MCC results of the \texttt{XGBoost} classifier with \texttt{ts-fresh} features across all temporal granularities. Similar to handcrafted features, performance declines with coarser resolutions. Figure~\ref{figure_results_precision_recall_large_home_zoomed} further illustrates the precision–recall trade-off for \textit{large\_home} and \textit{swimming\_pool}~\cite{Radovanovic25a}, providing a zoomed view that highlights the consistent performance drop with increasing granularity.

\newpage
\section{Additional Feature Importance Analysis}
\begin{figure*}[ht]
  \centering
  \includegraphics[width=\linewidth]{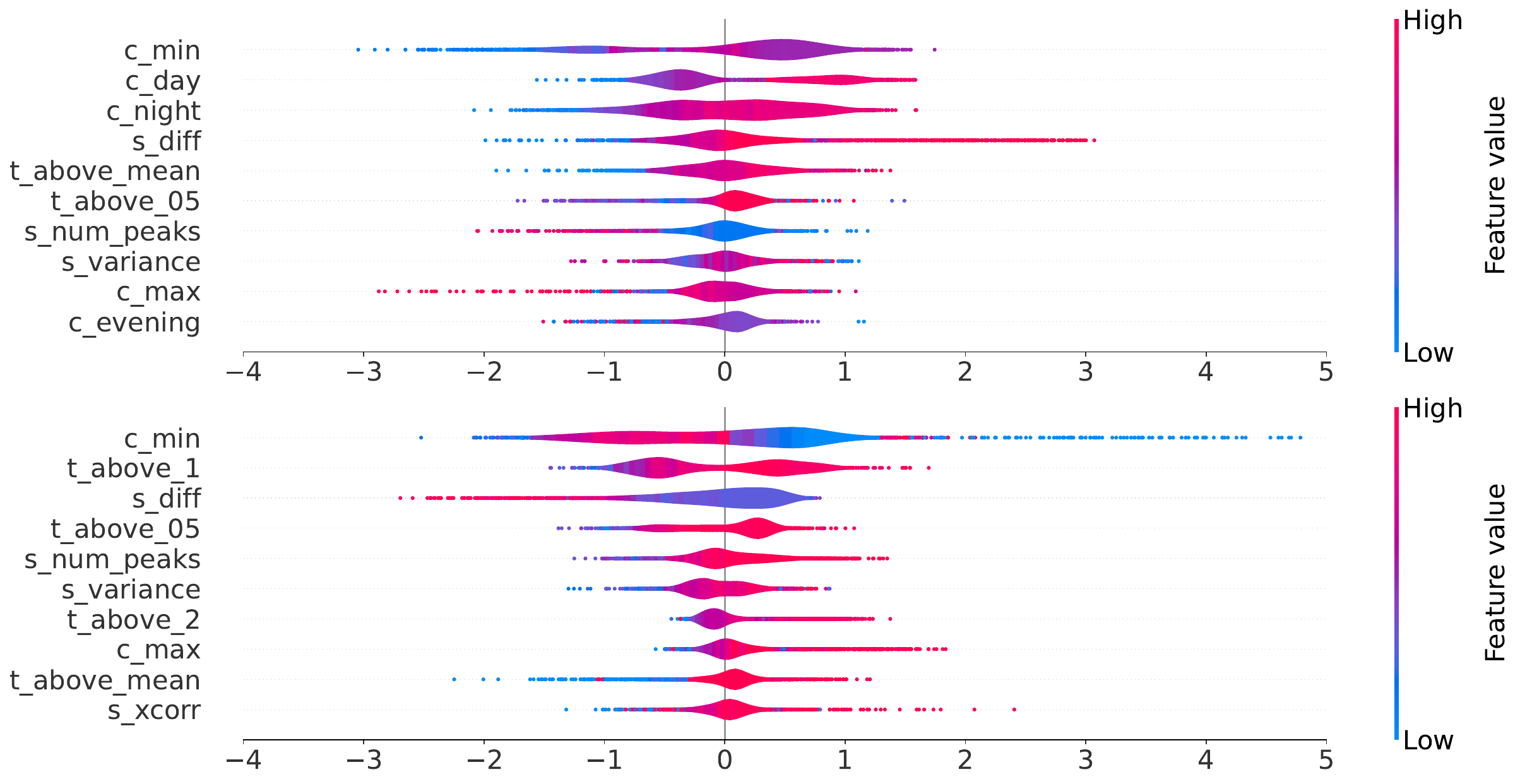}
  \caption{SHAP violin plots for \textit{large\_home} at 15 minutes and 4 hours.}
  \label{fig:FeatureImportance_Large_Home}
\end{figure*}

To further analyze the drivers of prediction, Figure~\ref{fig:FeatureImportance_Large_Home} illustrates violin plots of the ten most important handcrafted features for the attribute \textit{large\_home}. Results are shown at two temporal resolutions, 15 minutes and 4 hours. The y-axis lists the top features, while the x-axis represents SHAP values indicating their impact on the model output. Colors encode feature magnitudes, with red denoting high values and blue low values.  

\begin{figure*}[h!]
    \centering
    \includegraphics[width=\linewidth]{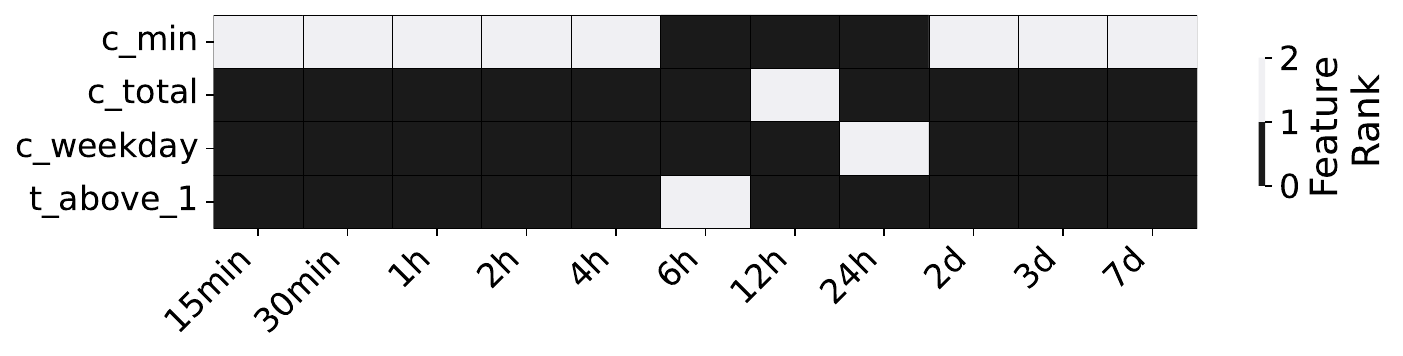}
    \caption{Most influential feature for \textit{large\_home} across granularities.}
  \label{fig:large_home_ranks_heatmap}
\end{figure*}

Complementing this view, Figure~\ref{fig:large_home_ranks_heatmap} shows how the most influential handcrafted feature shifts across all temporal resolutions. Each column corresponds to a granularity, and the lightest cell marks the dominant feature at that resolution. This highlights the relative stability of baseline consumption indicators such as \textit{c\_min}, contrasted with occasional shifts to other aggregates at specific granularities.  

\label{sec:Appendix}

\end{document}